\documentclass[letterpaper, 10 pt, conference]{ieeeconf}  

\IEEEoverridecommandlockouts                              

\usepackage{graphics} 
\usepackage{epsfig} 
\usepackage{mathptmx} 
\usepackage{times} 
\usepackage{amsmath} 
\usepackage{amssymb}  
\usepackage{subfigure}
\usepackage{multirow}
\usepackage{subcaption}
\usepackage{float}
\usepackage{siunitx}
\usepackage{ctable}
\usepackage{cuted}
\usepackage{colortbl}
\usepackage{algorithm}
\usepackage{algpseudocode}
\usepackage{titlesec}

\usepackage{soul}

\usepackage[numbers,sort&compress]{natbib}

\usepackage{eucal} 

\usepackage[colorlinks,linkcolor=blue,citecolor=blue]{hyperref}
\title{\LARGE \bf
Embodiment‑Aware Generalist Specialist Distillation for Unified Humanoid Whole-Body Control
}
\DeclareMathAlphabet{\mathcal}{OMS}{cmsy}{m}{n}

\author{
    Quanquan Peng\textsuperscript{1*}, Yunfeng Lin\textsuperscript{1*}, Yufei Xue\textsuperscript{1}, Jiangmiao Pang\textsuperscript{2}, Weinan Zhang\textsuperscript{1 2\dag} \\

{$^{1}$Shanghai Jiao Tong University, $^{2}$Shanghai AI Lab}

\thanks{* denotes equal contribution  \dag \ denotes corresponding author}
}
\begin{document}

\maketitle
\thispagestyle{empty}
\pagestyle{empty}

\begin{strip}
\vspace{-17mm}
\centering
\includegraphics[width=\textwidth]{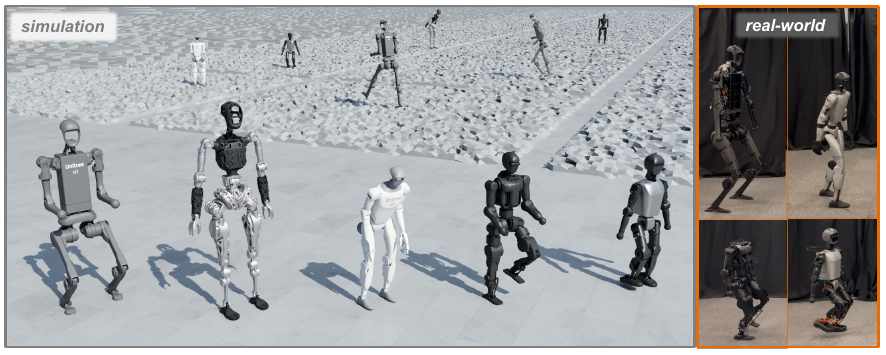}
\captionof{figure}{In this work, we propose a distillation framework that yields a \emph{single} whole-body controller that runs on heterogeneous humanoids. Through cyclic forking of embodiment-specific specialists and distillation back into the generalist, the resulting policy executes rich commands (walking, squatting, and leaning) in simulation and the real world, outperforming baselines in tracking accuracy and robustness.}\label{fig:teaser}
\end{strip}

\begin{abstract}
Humanoid Whole‑Body Controllers trained with reinforcement learning~(RL) have recently achieved remarkable performance, yet many target a single robot embodiment. Variations in dynamics, degrees of freedom~(DoFs), and kinematic topology still hinder a single policy from commanding diverse humanoids. Moreover, obtaining a generalist policy that not only transfers across embodiments but also supports richer behaviors---beyond simple walking to squatting, leaning---remains especially challenging.
In this work, we tackle these obstacles by introducing \textsc{EAGLE}, an iterative generalist-specialist distillation framework that produces a single unified policy that controls multiple heterogeneous humanoids without per‑robot reward tuning. During each cycle, embodiment‑specific specialists are forked from the current generalist, refined on their respective robots, and new skills are distilled back into the generalist by training on the pooled embodiment set. Repeating this loop until performance convergence produces a robust Whole-Body Controller validated on robots such as Unitree H1, G1, and Fourier N1.
We conducted experiments on five different robots in simulation and four in real-world settings. Through quantitative evaluations, \textsc{EAGLE} achieves high tracking accuracy and robustness compared to other methods, marking a step toward scalable, fleet‑level humanoid control.

\end{abstract}

\section{Introduction}
Training one policy that can transfer across different robot platforms is emerging as a key challenge in robotics research. In manipulation, for example, large-scale visuomotor models trained on cross-domain datasets have been shown to generalize to different manipulators in various tasks~\cite{o2024open,black2024pi_0}, while similar results also appear in navigation~\cite{zeng2024poliformer, liu2025compass}. These successes suggest that cross‑embodiment learning is feasible and promising.

For humanoid whole‑body control (WBC), however, the situation is different. Large‑scale, data‑hungry paradigms that work in manipulation, where human operators can teleoperate robot arms~\cite{cheng2024open, luo2024human, fang2024airexo} to collect hundreds of demonstrations for training, do not easily translate to legged locomotion, since a robot without an existing controller simply cannot be controlled via teleoperation. Therefore, this imitation-learning pipeline stalls at the data‑collection stage. As an alternative, reinforcement learning (RL) algorithms~\cite{schulman2017proximal, haarnoja2018soft} have propelled WBC to impressive heights, allowing robots to run, hop, dance, and manipulate~\cite{ji2024exbody2,xue2025unified,he2024omnih2o,he2024hover}. Yet almost every milestone is tied to one specific embodiment. Discrepancies in hardware attributes across dynamics, kinematics, and morphology block a single policy from transferring directly; each new robot typically forces the entire training and reward‑tuning pipeline to start over, slowing down deployment.
Recent studies have explored bridging this gap with diffusion‑based motion priors~\cite{yang2025multi} or by massively randomizing URDF parameters during training~\cite{bohlinger2024one, ai2025towards, liu2025locoformer}. However, these methods are still limited to low‑dimensional velocity commands, leaving other behaviors such as torso pitching beyond reach, and importantly, they have not yet been validated on diverse real humanoid hardware.

Naturally, a question was raised: \textit{Can one policy control multiple humanoids while still supporting diverse whole‑body commands?} In this work, we propose \textbf{EAGLE}, an embodiment‑aware generalist-specialist distillation framework for unified humanoid whole‑body control. EAGLE couples an iterative generalist-specialist distillation loop with a unified, high‑dimensional command interface. The loop begins with a generalist trained in a simulator that pools multiple different humanoid models (Unitree H1, Unitree G1, Booster T1, Fourier N1, and PNDbotics Adam); from this policy, we spawn embodiment‑specific specialists, fine‑tune them on their respective robots, and then distill their new skills back into the generalist in a DAgger-based~\cite{ross2011reduction} way, repeating the cycle until the generalist policy converges. Built on the HugWBC~\cite{xue2025unified}, our command vector encodes base linear and angular velocities, body pitch, and base height, enabling behaviors far richer than pure walking---such as leaning and squatting---without any per robot-specific reward tuning.

In summary, our key contributions are as follows:
\begin{itemize}
    \item We introduce an embodiment‑aware generalist-specialist distillation loop that unifies whole‑body control across heterogeneous humanoids without per-robot reward tuning.
    \item We deploy a unified high‑dimensional command interface, which allows one policy to perform squatting, leaning, and base velocity tracking, capabilities that previous approaches cannot support.
    \item We conduct extensive experiments on five humanoid robots in simulation and four in the real world, showing that \textsc{EAGLE} achieves higher command‑tracking accuracy and superior performance compared with baseline methods.
\end{itemize}

\section{Related Works}
\textbf{Cross-embodiment learning.} Training a model that is able to transfer across different robot platforms can be promising~\cite{yu2017preparing, ying2024peac, yang2024pushing}. Many works have attempted to address this issue by training on diverse datasets~\cite{o2024open, xu2023xskill, doshi2024scaling, black2024pi_0}, finding a unified action representation~\cite{unigrasp2020shao}, or domain randomization and adaptation~\cite{tobin2017domain}. Meanwhile, humanoid robots have more DoFs and distinct morphological structures, making the task of finding a general controller more challenging. Some RL-based works tried to adopt diffusion models along with a residual term~\cite{yang2025multi} or by training Transformers with a large pool of URDF randomized robots~\cite{bohlinger2024one, ai2025towards, liu2025locoformer}. Though substantial progress has been made, most methods still support only low-dimensional velocity commands, and other diverse behaviors remain underexplored. Moreover, these approaches remain limited to simulation studies and do not demonstrate transfer to multiple physical humanoid robots, which is the central focus of our work.

\textbf{Learning-based humanoid control.} Recent learning-based methods combine massively parallel simulation and reinforcement learning to achieve robust locomotion, recovery, and terrain traversal~\cite{kumar2021rma, cheng2023extreme, hwangbo2019learning}. Building on these advances, recent work on humanoids has spanned real-world whole-body control~\cite{he2024hover, he2024omnih2o, fu2024humanplus}, teleoperation interfaces~\cite{ze2025twist, cheng2024open}, and improved robustness under disturbances~\cite{huang2025learning}. Meanwhile, HugWBC~\cite{xue2025unified} introduces a unified, high-dimensional command space that enables a single policy to easily switch between walking, standing, and hopping, providing a versatile backbone for whole-body control. We adopt this framework and focus on closing the cross-embodiment gap.

\textbf{Policy distillation.} Many works study transferring knowledge from teacher to deployable student models. Classical knowledge distillation compresses a network into a smaller policy via soft targets~\cite{hinton2015distilling, gou2021knowledge, park2019relational}. For multi-task RL, Distral distills task policies into a shared policy to aid transfer~\cite{teh2017distral}, and Kickstarting warm-starts learners from a fixed teacher with a decaying KL~\cite{schmitt2018kickstarting}. DAgger aggregates on-policy states with teacher relabels to reduce covariate shift~\cite{ross2011reduction}. In robotics, teacher-student pipelines have proved effective for legged locomotion~\cite{miki2022learning, kumar2021rma}. Building on these ideas, we adopt iterative generalist-specialist distillation, producing a generalist policy that scales across heterogeneous humanoids while preserving rich whole-body commands.

\section{Preliminaries}
\label{sec:prelim}
We consider $N$ humanoid robots with heterogeneous morphologies.
For robot $i\!\in\!\{1,\dots,N\}$, we model the control problem as a
Markov Decision Process (MDP):
$\mathcal M_i=\langle \mathcal S_i,\mathcal A_i,\mathcal P_i,\mathcal R_i, H\rangle$, where
$x_t^i\!\in\!\mathcal S_i, a_t^i\!\in\!\mathcal A_i$ is the robot state and action at time step $t$,
$\mathcal P_i$ is the state-action transition dynamics,
$\mathcal R_i$ is the reward function,
and $H$ is the episode horizon.
Formally, our goal is to get a generalist policy $\pi_g$ such that, given command vector $c_t$, it can generate $a_t \sim \pi_g(o_t)$ to maximize the objective function: 
\begin{equation}
\frac{1}{N}\sum_{i=1}^{N}
\;
\mathbb{E}_{\tau_i\sim\pi_g}
\!\Bigl[
\sum_{t=0}^{H}
\gamma^{\,t}\,
\mathcal{R}_i\!\bigl(s_t^{\,i},c_t,a_t^{\,i}\bigr)
\Bigr],
\label{eq:cross_obj}
\end{equation}
where $\gamma\!\in\!(0,1)$ is the reward discount factor, $o_t$ is the $(s_t, c_t)$ pair, and  
$\tau_i=\{(s^{\,i}, {s^i}',a^{\,i}, r^i)_{0:H}\}$ denotes a rollout on embodiment $i$.


\section{Methodology}

In this section, we introduce \textsc{EAGLE}, our cross-embodiment training framework. 
Sec.~\ref{subsec:cmd} describes how we build a unified, high-dimensional command and observation space for humanoid control for diverse robot behaviors.  
Because the robots differ in their DoF counts and kinematic layouts, 
Sec.~\ref{subsec:align} explains how we align heterogeneous observation-action vectors 
, enabling the neural network to exploit morphological information while handling all embodiments.
Sec.~\ref{subsec:reward} then outlines the reward-function design used for RL training.  
Finally, Sec.~\ref{subsec:distill} details the iterative generalist--specialist distillation loop that transfers embodiment-specific skills back into a shared policy.

\begin{figure*}[tbh]
  \centering
  \includegraphics[width=0.9\textwidth]{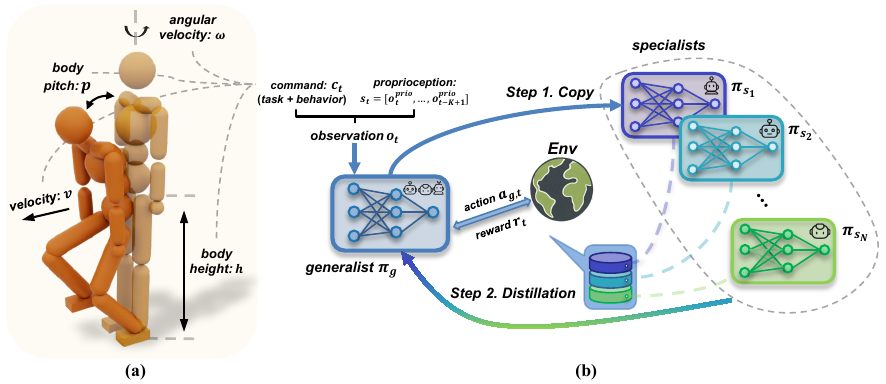}
    \caption{\textbf{Method Overview.} \textbf{(a)} Unified command interface. The command vector
    $\mathbf c_t$ comprises \emph{task} commands $\mathbf v_t$ (linear velocities $v_x,v_y$, angular velocity $\omega$) and \emph{behavior} commands $\mathbf b_t$ (base height $h$, body pitch $p$); together with a window of robot proprioception $s_t$, they form the observation $o_t$. \textbf{(b)} Generalist-specialist distillation framework. Each round \emph{copies} $\pi_g$ to $N$ specialists $\{\pi_{s_i}\}$ for per-robot fine-tuning, then \emph{distills} back by running $\pi_g$, relabeling actions with the corresponding specialist, and updating with the loss in Eq.~\ref{eqn:dagger-loss}. Repeating this loop yields a single controller that scales across embodiments while retaining rich whole-body commands.}
  \label{fig:overview}
\end{figure*}

\subsection{Unified Command and Observation Space.}
\label{subsec:cmd}
\subsubsection{Command Design}
To make the robot support richer behaviors, not limited to walking, we should design a unified high-dimensional command space. Inspired by the generalized command interface of HugWBC~\cite{xue2025unified}, we design the command vector (shown in Fig.~\ref{fig:overview}(a))
\begin{equation}
  \mathbf c_t \;=\;
  \bigl[\,v_{x},\; v_{y},\; \omega,\; h,\; p\bigr]^{\!\top}
  \in\mathbb{R}^{5},
  \label{eq:cmd}
\end{equation}
where \(v_x\) and \(v_y\) are desired base-frame linear velocities
(\si{m/s}), \(\omega\) is the angular velocity (\si{rad/s}), \(h\) is the
base height offset (\si{m}), and \(p\) is the body-pitch angle
(\si{rad}).  

The first three components $\mathbf v_t=[v_x,v_y,\omega]^\top$,
specify \emph{task command}: they define where the robot should go.
The last two $\mathbf b_t=[h,p]^\top$, modulate
\emph{behavior command}, let the robot do behaviors like squatting, leaning, or standing.
Separating $\mathbf v_t$ and $\mathbf b_t$ allows a single policy to realize
a rich repertoire of movements beyond simply walking. 

\subsubsection{Embodiment-aware Observation}
As for the policy observation, the actor receives a chunk of proprioception $o^{prio}$
(joint positions $q$, joint velocities $\dot q$, base angular velocity $\omega_{base}$, and projected gravity $\mathbf g^{\text{proj}}$ w.r.t. the robot's base frame) of $K=5$ frames: $s_t=[o^{prio}_t,\cdots, o^{prio}_{t-K+1}]$.
To help the robot learn foot lifting that follows a gait pattern, we add a gait clock function $\sin(2\pi \phi)$ in observation:
\begin{equation}
  \begin{aligned}
    \phi_{1,t+1} &= \phi_{1,t} + f \mathrm d t,\\
    \phi_{2,t+1} &= \phi_{1,t+1} + \varphi,
  \end{aligned}
  \label{eq:clock}
\end{equation}
where \(f \in \mathbb R \) is the command gait frequency
(\si{Hz}) and \(\varphi\in[0,1)\) is the phase 
offset between the two legs.
Moreover, we adopt the asymmetric actor-critic paradigm~\cite{choi2023learning}, where the critic receives additional privileged information: the base linear velocity, height error,
foot clearance and contact forces,  link collision state, terrain friction coefficient, and height scan samples.

To let the network reason about the differences in robot morphology, we additionally provide \emph{embodiment-aware observation} $o_{ea} \in \mathbb R^{m\times 10}$ as privileged information for the critic and task the actor to estimate it. Here $m$ is the number of predicted rigid bodies, and in this work, we pick the torso and both feet ($m=3$) to preserve locomotion-critical signals instead of upper-body links. For each rigid body,  $o_{ea}$ contains information of the corresponding mass (in $\mathbb R$), CoM position (in $\mathbb R^3$), and inertia matrix (in $\mathbb R^6$). Exp.~\ref{exp:latent-vec} shows that this helps the model to distinguish across robots and their locomotion patterns.

\subsection{Embodiment Alignment.}
\label{subsec:align}
As mentioned above, different humanoids expose distinct observation and action spaces, which prevents a single network from sharing weights across embodiments. We resolve this by embedding all robots into \emph{unified} observation-action spaces via zero padding and fixed index mappings. Specifically, we choose a unified action length $D_a = 32$, covering
lower limbs (6 DoFs per leg with hip RPY, knee P, ankle RP),
the waist (RPY), the head (RPY), and upper limbs (7 DoFs per arm with shoulder RPY, elbow P, wrist RPY), where RPY denotes roll/pitch/yaw.
For robot $i$, its native action $a^{i}_t\!\in\!\mathbb{R}^{n_i}$ is embedded as
\begin{equation}
\label{eq:embed-action}
\tilde a^{i}_t
\;=\;
P_m
\!\begin{bmatrix}
a^{i}_t\\[2pt]
\mathbf{0}
\end{bmatrix}
\;\in\;\mathbb{R}^{D_a},
\end{equation}
where $\mathbf{0}\in \mathbb R^{D_a-n_i}$ is a zero vector and $P_m\in\{0,1\}^{D_a\times D_a}$ is a robot-specific permutation matrix that maps each joint to a fixed global index (e.g., \texttt{Right-Hip-Pitch} joint $\rightarrow 0$, \texttt{Left-Hip-Pitch} joint $\rightarrow 1$, ...). 
At execution time, the environment wrapper applies the inverse mapping to recover native actions:
\begin{equation}
\label{eq:inverse-map}
a^{i}_t \;=\; S_i\, \tilde a^{i}_t,
\quad \text{where }
S_i \in \{0,1\}^{n_i\times D_a},\ \ S_i P_i
\begin{bmatrix}
I_{n_i} & \mathbf{0}
\end{bmatrix}^\top\!= I_{n_i}.
\end{equation}
We preserve robot-specific parameters, such as the swing height of both feet $\ell_{\mathrm{target},j} (j\in \{0, 1\})$, as well as joint stiffness and damping parameters, as these will significantly impact the training phase. 

Moreover, to train the generalist policy $\pi_g$, we will also spawn different humanoids together in parallel within the same training environment with equal probability. As for how to align them into the reward function $\mathcal{R}_i\!\bigl(\cdot \bigr)$, please see Sec.~\ref{subsec:reward} for details.

\subsection{Reward Function Design.}
\label{subsec:reward}
We adopt PPO~\cite{schulman2017proximal} algorithm with asymmetric actor-critic paradigm (see Sec.~\ref{subsec:cmd}) as our Reinforcement Learning algorithm. The per-step reward is decomposed into a task term $r_t^{\text{task}}$, a behavior term $r_t^{\text{beh}}$, and a regularization term $r_t^{\text{reg}}$:
\begin{equation}
  r_t \;=\; \omega_1\, r_t^{\text{task}} \;+\; \omega_2\, r_t^{\text{beh}} \;+\; \omega_3\, r_t^{\text{reg}}.
\end{equation}
Here, $r_t^{\text{task}}$ encourages tracking of the task command $\mathbf v_t$, $r_t^{\text{beh}}$ encourages tracking of the behavior command $\mathbf b_t$, and $r_t^{\text{reg}}$ penalizes failures and undesirable actions (e.g., falls, excessive torques, abrupt changes). For most of the reward, we inherit them from HugWBC~\cite{xue2025unified}. See Tab.~\ref{tab:reward} for details.

\begin{table}[t]
  \centering
  \caption{\textbf{Training reward terms} (for simplicity, note that \(\rho_{\kappa}(x)=\exp(-x/\kappa)\)). For the definition of $C(\phi)$ function, please refer to HugWBC~\cite{xue2025unified}.}
  \small
  \setlength{\tabcolsep}{3pt}
  \renewcommand{\arraystretch}{1.15}
  \begin{tabularx}{\columnwidth}{l >{\centering\arraybackslash}X r}
    \toprule
    \textbf{Term} & \textbf{Definition} & \textbf{Coefficient}\\
    \midrule
    \multicolumn{3}{c}{\emph{Task Reward}: $r_t^{\text{task}}$ }\\
    \midrule
    Lin. vel. tracking     & \(\rho_{0.25}\!\big(\|\mathbf v_{xy}^{\mathrm{target}}-\mathbf v_{xy}\|^2\big)\) & \(2\)\\
    Ang. vel. tracking      & \(\rho_{0.25}\!\big(\|\omega_z^{\mathrm{target}}-\omega_z\|^2\big)\)             & \(2.5\)\\
    \midrule
    \multicolumn{3}{c}{\emph{Behavior Reward}: $r_t^{\text{beh}}$}\\
    \midrule
    Base height             & \(\|h^{\mathrm{target}}-h\|^2\)                                                  & \(-60\)\\
    Body pitch             & \(\|p^{\mathrm{target}}-p\|^2\)                                                  & \(-1\)\\
    Foot-swing position & {\scalebox{0.92}{$\displaystyle \sum_{j}\!\bigl(1{-}C(\phi_j)\bigr)\,\bigl\|\ell_{\mathrm{target},j} - \ell_{\mathrm{foot},j}\bigr\|^{2}$}} & $-30$\\
    Contact vel. & {\scalebox{0.92}{$\displaystyle \!\sum_{j} C(\phi_j)\Bigl[\,1-\rho_5 (\|\mathbf v^{\mathrm{foot}}_{xy,j}\|^{2})\Bigr]$}}
& $-1$\\
    Contact force & {\scalebox{0.92}{$\displaystyle \!\sum_{j} \bigl(1-C(\phi_j)\bigr)\Bigl[\,1-\rho_{50} (\|\mathbf f^{\mathrm{foot}}_{xy,j}\|^{2})\Bigr]$}}
& $-1$\\

    \midrule
    \multicolumn{3}{c}{\emph{Regularization Reward}: $r_t^{\text{reg}}$}\\
    \midrule
    Roll-pitch Ang. vel.   & \(\|\boldsymbol{\omega}_{xy}\|^2\)                                         & \(-0.1\)\\
    Vertical speed         & \(\|v_z\|^2\)                                                                 & \(-2\)\\
    Foot slip              & \(1-\sum_j \rho_{1}(\|\mathbf v^{\mathrm{foot}, j}_{xy}\|^2)\)                 & \(-0.1\)\\
    Action rate            & \(\|a_t-a_{t-1}\|^2\)                                                        & \(-2\!\times\!10^{-3}\)\\
    Action smoothness      & \(\|a_{t-2}-2a_{t-1}+a_t\|^2\)                                              & \(-2\!\times\!10^{-3}\)\\
    Joint torque           & \(\|\boldsymbol{\tau}\|^2\)                                                 & \(-1\!\times\!10^{-5}\)\\
    Joint acceleration     & \(\|\ddot{\mathbf q}\|^2\)                                                  & \(-5\!\times\!10^{-8}\)\\
    Upper-joint deviation  & \(\|\mathbf q^{\mathrm{default}}_{\mathrm{upper}}-\mathbf q_{\mathrm{upper}}\|^2\) & \(-5\)\\
    Hip-joint deviation  & \(\|\mathbf q^{\mathrm{default}}_{\mathrm{hip}}-\mathbf q_{\mathrm{hip}}\|^2\)            & \(-0.4\)\\
    Base Orientation       & $\|\mathbf g^{\text{proj}}_{xy}\|^{2}$                                       & \(-5\) \\
    \bottomrule
  \end{tabularx}

    \label{tab:reward}
    \vspace{-12pt}
\end{table}

Since we train multiple robots jointly, reward terms must be aligned across embodiments. In practice, we keep the weights shared across robots and adjust only term-specific targets or scales to account for morphology. For example, in the base-height tracking term $\|h^{\text{target}} - h\|^2$, the target $h^{\text{target}}$ is set per robot (e.g., its nominal base height).

\begin{algorithm}[t]
	\caption{Overall Process}
	\label{alg:overall}
	\begin{algorithmic}[1]
		\Require Number of different robots: $N$, Rollout buffer: $\mathcal D$;  
		\State Train generalist policy $\pi_g$ on $N$ robots; \Comment{see Sec.~\ref{subsec:align} for details}
		\While{$\pi_g$ doesn't converge}
            \State For $i\!\in\!\{1,\dots,N\}$, $\pi_{s_i} \leftarrow \pi_g$; \Comment{initialize model by copying weight}
            \State Fine-tune each $\pi_{s_i}$ on $i$-th robot;
            \For{$k = 1$ \textbf{to} $T$ epochs}
                \State $\mathcal D_{\pi_g} \leftarrow \{(s, s', a_{g}, r)\}$; \Comment{collect parallel env trajectories}
                \State Relabel $\mathcal D_{\pi_g}$ with specialist action $a_s$ from $\pi_{s_{1:N}}$;
                \State Update $\pi_{g}$ by Eq.~\ref{eqn:dagger-loss};
            \EndFor
        \EndWhile
		\State \Return $\pi_g$;
	\end{algorithmic}
\end{algorithm}

\subsection{Distillation Loop Overview.}
\label{subsec:distill}
So far, the alignment scheme in Sec.~\ref{subsec:align} enables a PPO policy $\pi_g$ to control different robots, but its performance soon saturates at a mediocre level.  To further improve it, we introduce an iterative \emph{generalist-specialist distillation loop} (shown in Fig.~\ref{fig:overview}(b)), that alternates between \emph{specialize} phase and \emph{generalize} phase:

\begin{enumerate}
  \item \textbf{specialize.}  
        Copy the current generalist to $N$ robot-specific specialists
        $\{\pi_{s_i}\}$ and fine-tune each on its own robot only.
  \item \textbf{generalize.}  
        Collect trajectories of different robots by running the $\pi_g$. Then relabel actions proposed by the corresponding specialists, and distill $\pi_g$ with it.
\end{enumerate}

As for the distillation mentioned above, prior approaches such as standard DAgger only align the output action distributions:
\begin{equation}
    \mathcal L_a=\mathbb{E}_{\tau \sim \pi_g} \mathbb E_{o_t\sim \tau} \bigl(\pi_g(o_t)-\pi_s(o_t)\bigr)^2.
\end{equation}
However, since the generalist and specialist share the same model architectures, our framework introduces an additional representation-level alignment loss:
\begin{equation}
\label{eq:embed}
 \mathcal L_e=\mathbb{E}_{\tau \sim \pi_g} \mathbb E_{o_t\sim \tau} \bigl(e_{\pi_g}(s_t) - e_{\pi_s}(s_t)\bigr)^2,   
\end{equation}
where $e(\cdot)$ denotes the \emph{hidden feature} of the actor network by taking proprioception buffer $s_t$, so
$\mathcal{L}_e$ aligns the two policies in representation space rather than actions alone.
We show in Sec.~\ref{exp:latent-vec} that this loss term improves cross-embodiment training.
Moreover, we want to maintain the critic updated while also preserving the generalist model's exploration ability, instead of simply mimicking specialists' behavior. Therefore, we keep the PPO loss for RL exploration, and the total distillation loss becomes:
\begin{equation}
\label{eqn:dagger-loss}
    \mathcal{L}=\mathcal{L}_{\textit{PPO}}+\alpha \cdot \mathcal L_a + \beta \cdot \mathcal{L}_e,
\end{equation}
where $\alpha=0.02$ and $\beta=1$ are constant coefficients.
Algo.~\ref{alg:overall} summarizes the full procedure.

This two-way exchange lets the generalist improve steadily while each specialist begins the next round from a stronger baseline, boosting overall control quality and eliminating per-robot reward tuning or network redesign.  

\section{Experiments}
In our experiment, we evaluate our method on five different robots in simulation and four in the real world. We intend to answer the following questions:
\begin{itemize}
    \item \textbf{Q1}: How much does \textsc{EAGLE} improve command-tracking accuracy compared to other methods?
    \item \textbf{Q2}: How does each component of the generalist-specialist cycle contribute to final performance?
    \item \textbf{Q3}: What embodiment-aware representation does the \textsc{EAGLE} policy learn?
    \item \textbf{Q4}: How well does the \textsc{EAGLE} policy transfer to real robots in a zero-shot way? 
\end{itemize}

\textbf{Experiment Setup.} In this work, we test on five humanoid models: Unitree H1, Unitree G1, Booster T1, Fourier N1, and PNDbotics Adam, whose DoFs are 19, 29, 23, 23, 25, respectively. The simulation training is done in Isaac Gym~\cite{makoviychuk2021isaac} with 4096 parallel environments, each robot type being sampled with equal probability.

\textbf{Comparison Method.} For simplicity, we abbreviate the baseline methods as follows:
\begin{itemize}
    \item PPO: RL baseline policy trained on all robots together, \textit{i.e.}, the $\pi_g$ at line 1 of Algo.~\ref{alg:overall}.
    \item PPO \emph{w/o} EO: the same policy but trained without embodiment-aware observation (Sec.~\ref{subsec:cmd}) and the corresponding reconstruction loss. 
    \item COMPASS~\cite{liu2025compass}: We reimplement COMPASS, a three-stage cross-embodiment method that first imitates demonstrations to learn a world-model-conditioned base policy and then trains residual RL specialists and uses KL loss to distill them into a general policy. 
    \item Kickstarting~\cite{schmitt2018kickstarting}: 
    a distillation method in which \(\pi_g\) is updated by
        $
          \mathcal{L}
          =\mathcal{L}_{\textsc{PPO}}
          +\beta_{t}\,
           \mathrm{KL}\!\bigl(\pi_{s}\Vert\pi_g \bigr),
        $
        where \(\pi_s\) is the specialist policy and the coefficient
        \(\beta_{t}=\alpha exp(-\lambda t)\) decays exponentially.
    \item \textsc{EAGLE} (ours): our framework after a \emph{single} round of distillation, the same as Kickstarting.
    \item \textsc{EAGLE} \emph{w/} ID (ours): the full iterative distillation variant until the generalist's performance converges.
\end{itemize}
We also use the same model architecture and reward functions for fair comparison.

\subsection{Command Tracking Accuracy Analysis.}

\begin{table*}[t]
  \small
  \setlength{\tabcolsep}{4pt}
  \centering
  \caption{\textbf{Command-tracking errors} (lower is better). $E_{v_x}$ and $E_{v_y}$ are linear-velocity error in $\mathrm{m}/s$, $E_{\omega}$ is angular-velocity error in $\mathrm{rad}/s$, $E_h$ is base height error in $\mathrm{m}$, and $E_p$ is body-pitch error in $\mathrm{rad}$. Each tracking error is averaged over 2048 environments and 1000 steps. Our method maintains low errors across all robots compared to other cross-embodiment baselines.}
\begin{tabularx}{\textwidth}{l *{15}{>{\centering\arraybackslash}X}}
  \toprule
  \multirow{2}{*}{\shortstack{\textbf{Robot Type} \\ \\ Tracking Error ($\downarrow$)}}
    & \multicolumn{5}{c}{\textbf{Unitree H1}}
    & \multicolumn{5}{c}{\textbf{Unitree G1}}
    & \multicolumn{5}{c}{\textbf{PNDbotics Adam}} \\
  \cmidrule(lr){2-6}\cmidrule(lr){7-11}\cmidrule(lr){12-16}
    & $E_{v_x}$ & $E_{v_y}$ & $E_{\omega}$ & $E_h$ & $E_p$
    & $E_{v_x}$ & $E_{v_y}$ & $E_{\omega}$ & $E_h$ & $E_p$
    & $E_{v_x}$ & $E_{v_y}$ & $E_{\omega}$ & $E_h$ & $E_p$ \\ \midrule

  \rowcolor{gray!15}\multicolumn{16}{l}{(a) Ablation: embodiment-aware observation}\\\addlinespace[0.2em]
  PPO                 & 0.108 & 0.129 & 0.114 & \textbf{0.075} & 0.182
                      & 0.092 & 0.127 & 0.121 & 0.051 & \textbf{0.164}
                      & 0.328 & 0.107 & 0.235 & 0.051 & 0.204 \\

  PPO \emph{w/o} EO & 0.290 & 0.161 & 0.152 & 0.078 & 0.189
                      & 0.104 & 0.113 & 0.095 & 0.052 & 0.165
                      & 0.375 & 0.158 & 0.310 & 0.063 & 0.218 \\

  \rowcolor{gray!15}\multicolumn{16}{l}{(b) Ablation: baseline \& distillation method}\\\addlinespace[0.2em]
    COMPASS~\cite{liu2025compass}
                      & 0.725 & 0.390 & 0.476 & 0.085 & 0.273
                      & 0.171 & 0.119 & 0.124 & \textbf{0.035} & 0.169
                      & 0.763 & 0.371 & 0.339 & 0.045 & 0.233 \\
    Kickstarting~\cite{schmitt2018kickstarting}
                      & 0.323 & 0.141 & 0.122 & \textbf{0.075} & 0.182
                      & 0.117 & 0.118 & 0.075 & 0.050 & 0.167
                      & 0.399 & 0.086 & 0.141 & \textbf{0.041} & \textbf{0.192} \\
    \textsc{EAGLE} (ours)
                      & 0.061 & 0.117 & \textbf{0.054} & 0.076 & \textbf{0.169}
                      & \textbf{0.066} & 0.105 & 0.077 & 0.049 & \textbf{0.164}
                      & 0.182 & \textbf{0.084} & \textbf{0.115} & 0.050 & 0.196 \\
   \textsc{EAGLE} \emph{w/} ID (ours)
                      & \textbf{0.051} & \textbf{0.110} & \textbf{0.054} & 0.082 & 0.186
                      & 0.067 & \textbf{0.095} & \textbf{0.071} & 0.047 & 0.176
                      & \textbf{0.156} & 0.101 & 0.117 & 0.056 & 0.197 \\ \midrule 
\end{tabularx}

\vspace{0.8em}

\begin{minipage}{0.67\textwidth}
  \centering
  \begin{tabularx}{\linewidth}{l *{10}{>{\centering\arraybackslash}X}}
    \toprule
    \multirow{2}{*}{\shortstack{\textbf{Robot Type} \\ \\ Tracking Error ($\downarrow$)}}
      & \multicolumn{5}{c}{\textbf{Fourier N1}}
      & \multicolumn{5}{c}{\textbf{Booster T1}} \\
    \cmidrule(lr){2-6}\cmidrule(lr){7-11}
      & $E_{v_x}$ & $E_{v_y}$ & $E_{\omega}$ & $E_h$ & $E_p$
      & $E_{v_x}$ & $E_{v_y}$ & $E_{\omega}$ & $E_h$ & $E_p$ \\ \midrule

    \rowcolor{gray!15}\multicolumn{11}{l}{(a) Ablation: embodiment-aware observation}\\\addlinespace[0.2em]
    PPO                 & 0.124 & 0.141 & 0.107 & 0.073 & 0.165
                        & 0.147 & 0.203 & 0.105 & 0.047 & 0.156 \\

    PPO \emph{w/o} EO & 0.126 & 0.144 & 0.076 & 0.079 & 0.162
                        & 0.140 & 0.194 & 0.104 & 0.046 & 0.151 \\

    \rowcolor{gray!15}\multicolumn{11}{l}{(b) Ablation: baseline \& distillation method}\\\addlinespace[0.2em]
    COMPASS~\cite{liu2025compass}
                        & 0.060 & \textbf{0.117} & 0.137 & \textbf{0.056} & 0.160
                        & 0.831 & 0.307 & 0.179 & \textbf{0.044} & 0.151 \\
    Kickstarting~\cite{schmitt2018kickstarting}
                        & 0.147 & 0.142 & 0.130 & 0.077 & 0.168
                        & 0.761 & 0.190 & 0.100 & 0.047 & \textbf{0.149} \\
    \textsc{EAGLE} (ours) & 0.068 & 0.126 & 0.058 & 0.074 & \textbf{0.157}
                          & 0.109 & \textbf{0.187} & 0.045 & 0.045 & 0.152 \\
                          
    \textsc{EAGLE} \emph{w/} ID (ours)
                          & \textbf{0.056} & 0.135 & \textbf{0.056} & 0.074 & 0.168
                          & \textbf{0.090} & \textbf{0.187} & \textbf{0.041} & \textbf{0.044} & 0.159 \\ \midrule
  \end{tabularx}
\end{minipage}

  \label{tab:cmd-track}
  \vspace{-12pt} 
\end{table*}

\begin{table}[t]
  \small
  \setlength{\tabcolsep}{4pt}           
  \renewcommand{\arraystretch}{1.0}     
  \centering
  \caption{\textbf{Command ranges.} ``Initial'' and ``Finishing'' denote the RL curriculum learning stage (all in SI units).}
  \begin{tabular}{l c c c}
    \toprule
    \textbf{Group} & \textbf{Term} &
    \textbf{Initial Range} & \textbf{Finishing Range} \\
    \midrule
    \multirow{3}{*}{\shortstack{Task \\ Commands}}
      & $v_x$    & $[-0.3,\,0.6]$ & $[-0.6,\,1.2]$ \\
      & $v_y$    & $[-0.5,\,0.5]$ & $[-0.4,\,0.4]$ \\
      & $\omega$ & $[-0.5,\,0.5]$ & $[-1.0,\,1.0]$ \\
    \midrule
    \multirow{2}{*}{\shortstack{Behavior \\ Commands}}
      & $h$      & $[-0.3,\,0]$   & $[-0.3,\,0]$ \\
      & $p$      & $[-0.3,\,0.5]$ & $[-0.3,\,0.5]$ \\
    \bottomrule
  \end{tabular}
  \label{tab:cmd-range}
  \vspace{-12pt} 
\end{table}

To quantitatively understand how well our method compares to baseline methods, we refer to the command tracking error.

\textbf{Metric definition.} For each command dimension \(d\in\{v_x,v_y,\omega,h,p\}\)
we define the \emph{episode tracking error}:
\begin{equation}
  E_d \;=\;
  \mathbb{E}_{\tau \sim \pi}\!\left[
    \bigl|\,\hat c_t^{\,d}-c_t^{\,d}\bigr|
  \right],
  \label{eq:E_d}
\end{equation}
where \(c_t^{\,d}\) is the commanded target drawn from the predefined
range (Tab.~\ref{tab:cmd-range}), and
\(\hat c_t^{\,d}\) is the value read from the robot's proprioceptive state.  
We report five errors
\(E_{v_x},E_{v_y},E_{\omega},E_{h},E_{p}\) separately; lower values
indicate more accurate command tracking.

\textbf{Results.} Tab.~\ref{tab:cmd-track} summarizes the episode-averaged single command tracking errors.
We can see that compared to the PPO method, which was trained on all different robots together, both
\textsc{EAGLE} and \textsc{EAGLE} \emph{w/} ID surpass it on most metrics. 
As for the ablation on distillation method (Tab.~\ref{tab:cmd-track}(b)), compared with the KL-based Kickstarting baseline,
our single-round \textsc{EAGLE} already wins on 80\% robot–metric pairs and is never catastrophic,
whereas Kickstarting can become unstable on certain embodiment like G1, Adam and T1 (e.g. $E_{v_x}{=}0.761$ on T1 is five times higher than PPO).
The result confirms that the DAgger-style loss in Eq.~\ref{eqn:dagger-loss} is more robust across very different embodiments.

We also compare with the cross-embodiment baseline \textbf{COMPASS}~\cite{liu2025compass}. Although it achieves good results
on some robots like Fourier N1, it fails to generalize across all embodiments. In contrast, our method maintains low $E_{v_x}$ on all robots while keeping other metrics competitive or superior, demonstrating stable, single-policy control across heterogeneous humanoids.

\subsection{Ablation of the Distillation Loop.}

\begin{figure*}[t]
  \centering
  \includegraphics[height=3.55cm]{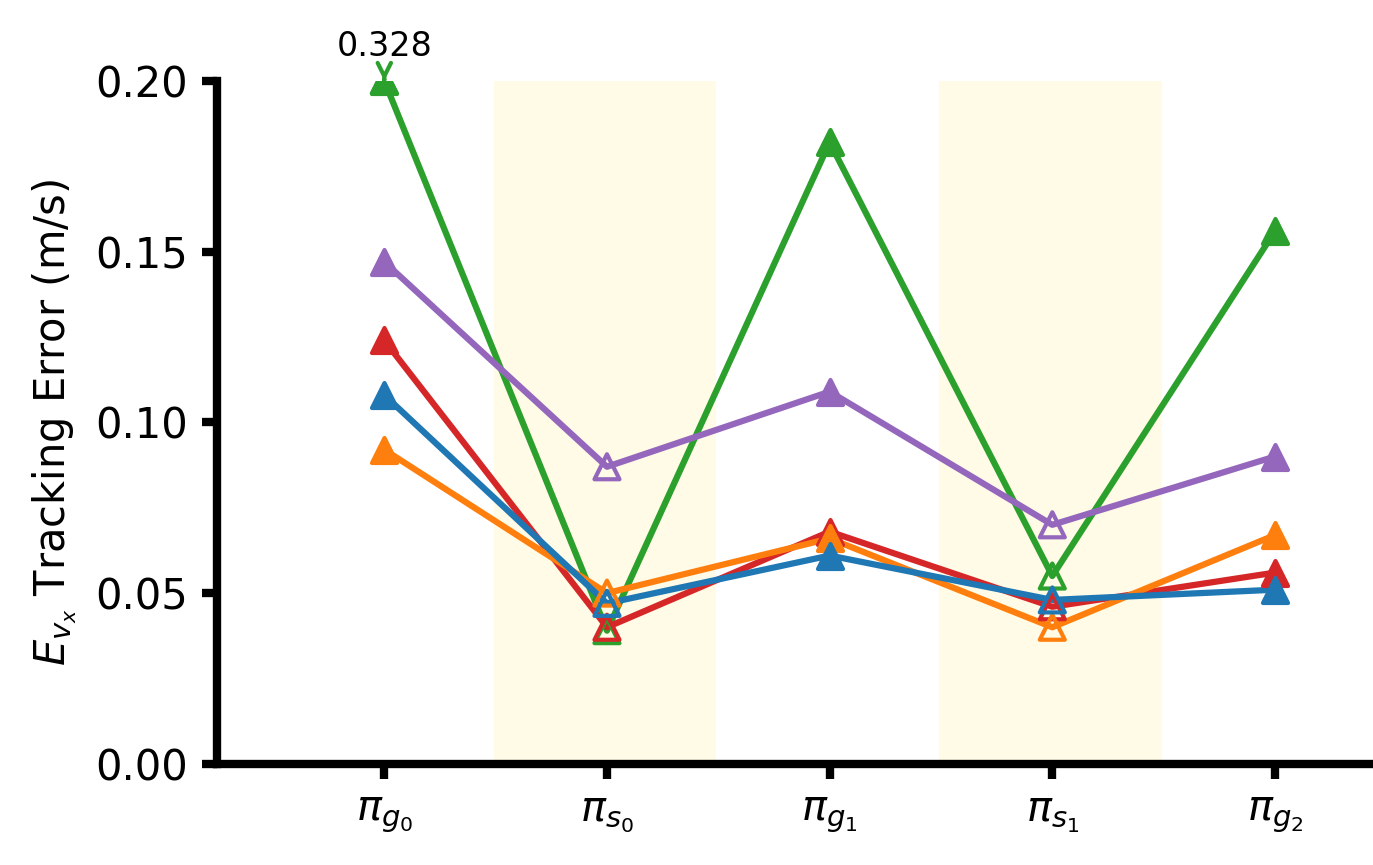}\hfill
  \includegraphics[height=3.55cm]{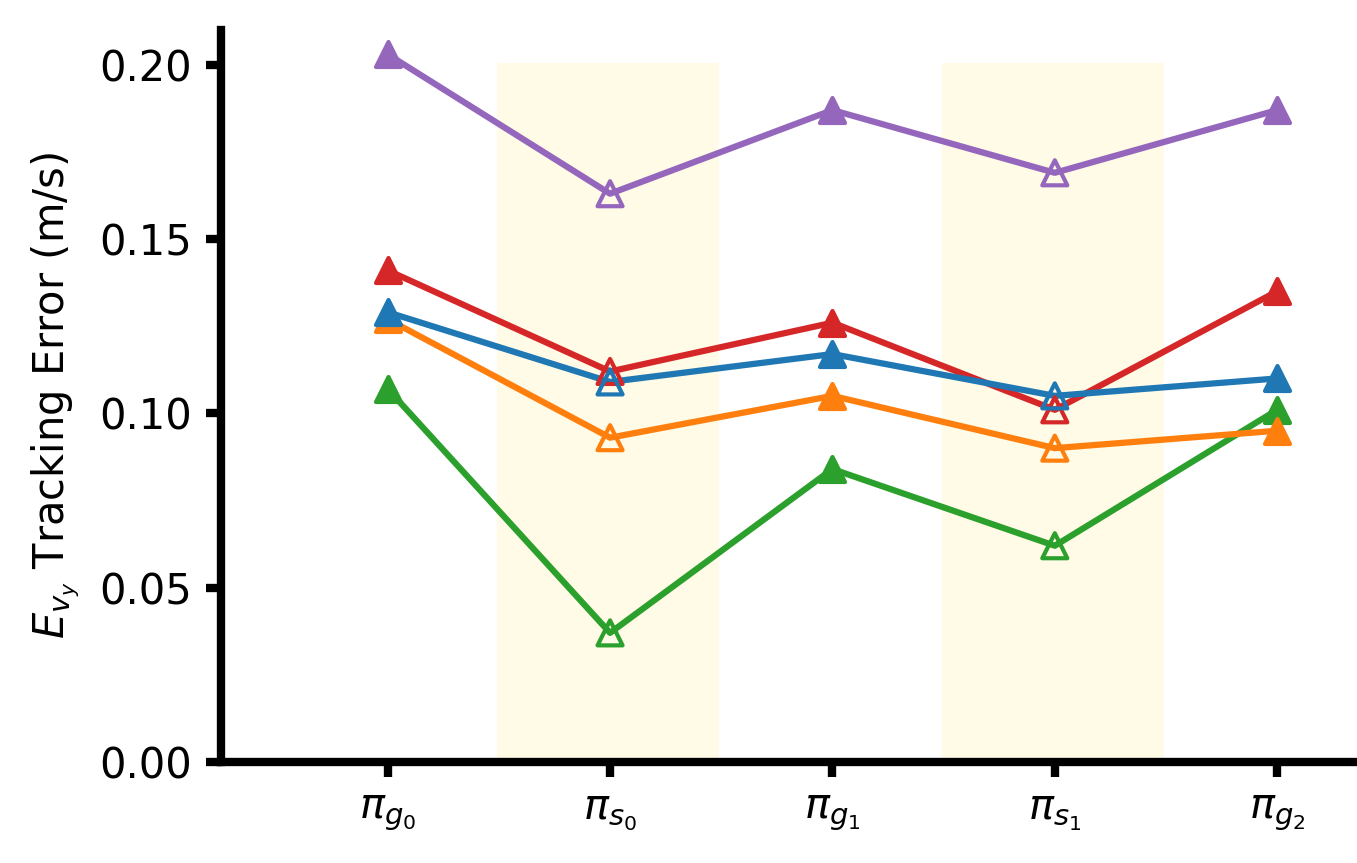}\hfill
  \includegraphics[height=3.55cm]{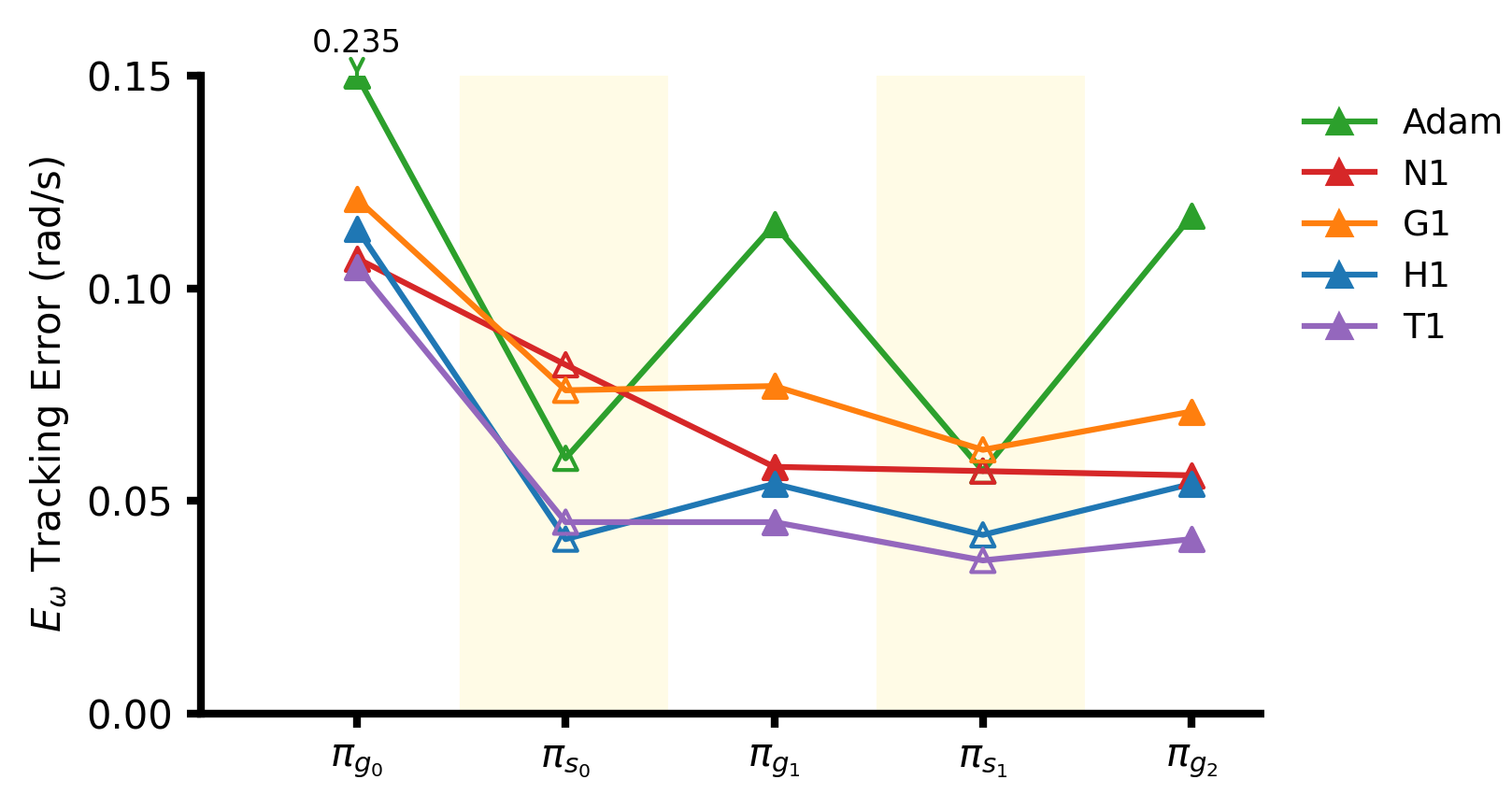}

  \caption{\textbf{Tracking-error curves for $E_{v_x}$, $E_{v_y}$, and $E_{\omega}$} (left to right). The \colorbox[HTML]{FFFBE6}{\scalebox{1.3}{$\vartriangle$}}, \scalebox{1.3}{$\blacktriangle$} represents the specialist and generalist policy respectively (as described at lines 2-9 of Algo.~\ref{alg:overall}). We can see the trend that, after each distillation cycle, both symbols move downward, showing that iterative distillation steadily improves tracking accuracy for specialists and generalists.}
  \label{fig:tracking-errors}
  \vspace{-12pt} 
\end{figure*}

\begin{table}[t]
  \small
  \setlength{\tabcolsep}{4pt}
  \centering
  \caption{\textbf{Unitree H1 command-tracking errors} (lower is better). After cross-embodiment training, \textsc{EAGLE}-generalist matches, and its robot-specific policy \textsc{EAGLE}-specialist surpasses a policy trained
exclusively on H1.}
  \begin{tabularx}{\linewidth}{l *{5}{>{\centering\arraybackslash}X}}
    \toprule
    \textbf{Tracking Error ($\downarrow$)}
      & $E_{v_x}$ & $E_{v_y}$ & $E_{\omega}$ & $E_h$ & $E_p$ \\
    \midrule

    single-robot PPO    & 0.067 & \textbf{0.101} & 0.052 & \textbf{0.072} & 0.181 \\
    \textsc{EAGLE}-generalist
                        & 0.061 & 0.117 & 0.054 & 0.076 & \textbf{0.169} \\
    \textsc{EAGLE}-specialist
                        & \textbf{0.048} & 0.105 & \textbf{0.042} & \textbf{0.072} & 0.174 \\
    \midrule
  \end{tabularx}
  
    \label{tab:unitreeh1}
    \vspace{-12pt} 
\end{table}

Furthermore, Tab.~\ref{tab:cmd-track} contrasts \textsc{EAGLE} after a \emph{single}
distillation round with its iterative variant.
For most task commands, \textsc{EAGLE} \emph{w/} ID achieves lower task-command errors compared to \textsc{EAGLE}, which wasn't trained after the iterative loop. 
Although a few behavior-command errors increase slightly, the overall
gains outweigh these minor regressions.  Fig.~\ref{fig:tracking-errors} plots the evolution of task-command errors across rounds and shows that,
after each distillation cycle, both the specialists (highlighted in yellow) and the generalists improve in most cases. 
This suggests that repeating the loop until convergence is more effective than a single distillation pass.

We additionally train \textit{single-robot} PPO policies that are exposed to only one robot type. As Tab.~\ref{tab:unitreeh1} shows, on robots such as H1, our generalist achieves performance comparable to these specialized baselines, while the corresponding \textsc{EAGLE}-specialist even surpasses them. 
This means cross-embodiment training has the potential to match, and sometimes exceed, policies tailored to a single embodiment.

\subsection{Analysis of Learned Representations.}
\label{exp:latent-vec}


\begin{figure}[tbh]
  \centering
  \includegraphics[width=1.05\columnwidth]{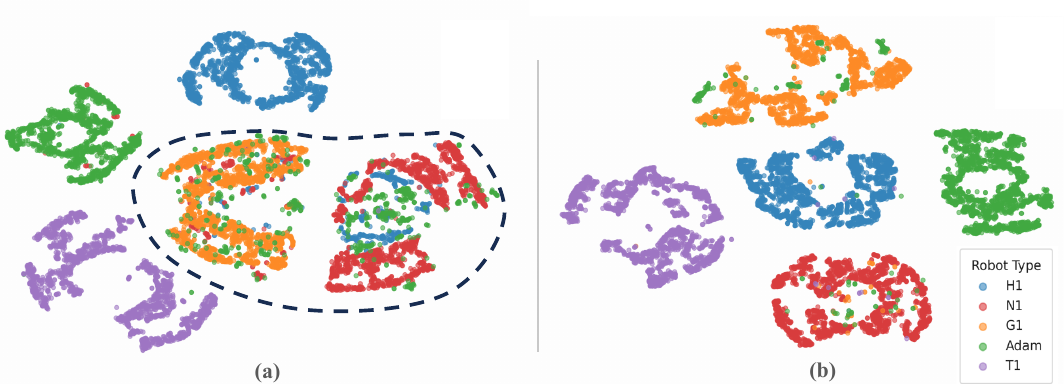}
  \caption{\textbf{t-SNE visualization of policy latent-space representation.} By commanding different robots to walk forward for certain timesteps, we can see \emph{w/o} embodiment-aware observation \textit{(left)}, the policy outputs latent vectors $e_\pi(\cdot)$ collapse into overlapping clusters. In contrast, our method \textit{(right)} is better separated, showing successful morphology-aware representation.}
  \label{fig:embed}
  
    \vspace{-12pt} 
\end{figure}

\begin{figure}[t]
  \centering
  \includegraphics[width=0.95\columnwidth]{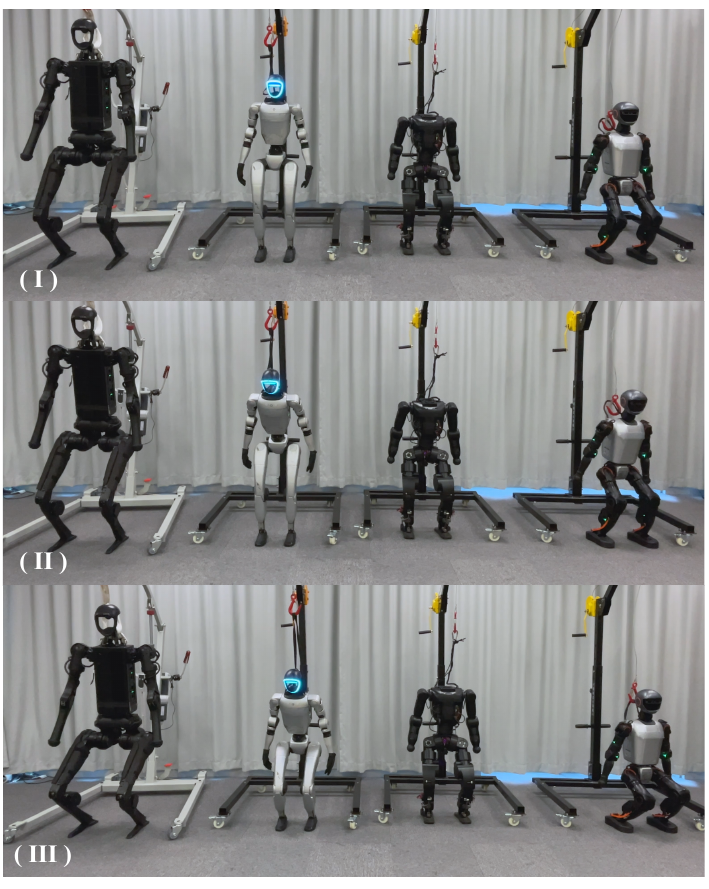} 
  \caption{\textbf{Real-world synchronous movement.} We also developed a shared low-level control framework to show our policy can let different robots execute synchronous movement, from leaning (II) to squatting (III). }
  \label{fig:real-exp2}
    \vspace{-12pt} 
\end{figure}

As we can see in Tab.~\ref{tab:cmd-track}(a), without the embodiment-aware observation, PPO \emph{w/o} EO's overall performance drops, and for some robots like H1, its $E_{v_x}$ tracking error increases significantly, showing that the policy fails to integrate different robot-specific knowledge together.

In Fig.~\ref{fig:embed}, we also use t-SNE to visualize the latent embedding $e_\pi(\cdot)$ (in Eq.~\ref{eq:embed}) by commanding the robot to walk forward. The \emph{left} panel depicts the PPO trained \emph{without} embodiment-aware observation: several robot types collapse into an
overlapping cluster (dashed ellipse), indicating that the encoder cannot
distinguish their dynamics. In contrast, the \emph{right} panel shows our policy, whose latent space
forms well-separated clusters for all five embodiments, confirming that the embodiment cue helps the network learn morphology-specific
representations.

\subsection{Zero-shot Sim2real Performance.} 
\label{exp:real}

\begin{figure}[t]
  \centering
  \includegraphics[width=0.85\columnwidth]{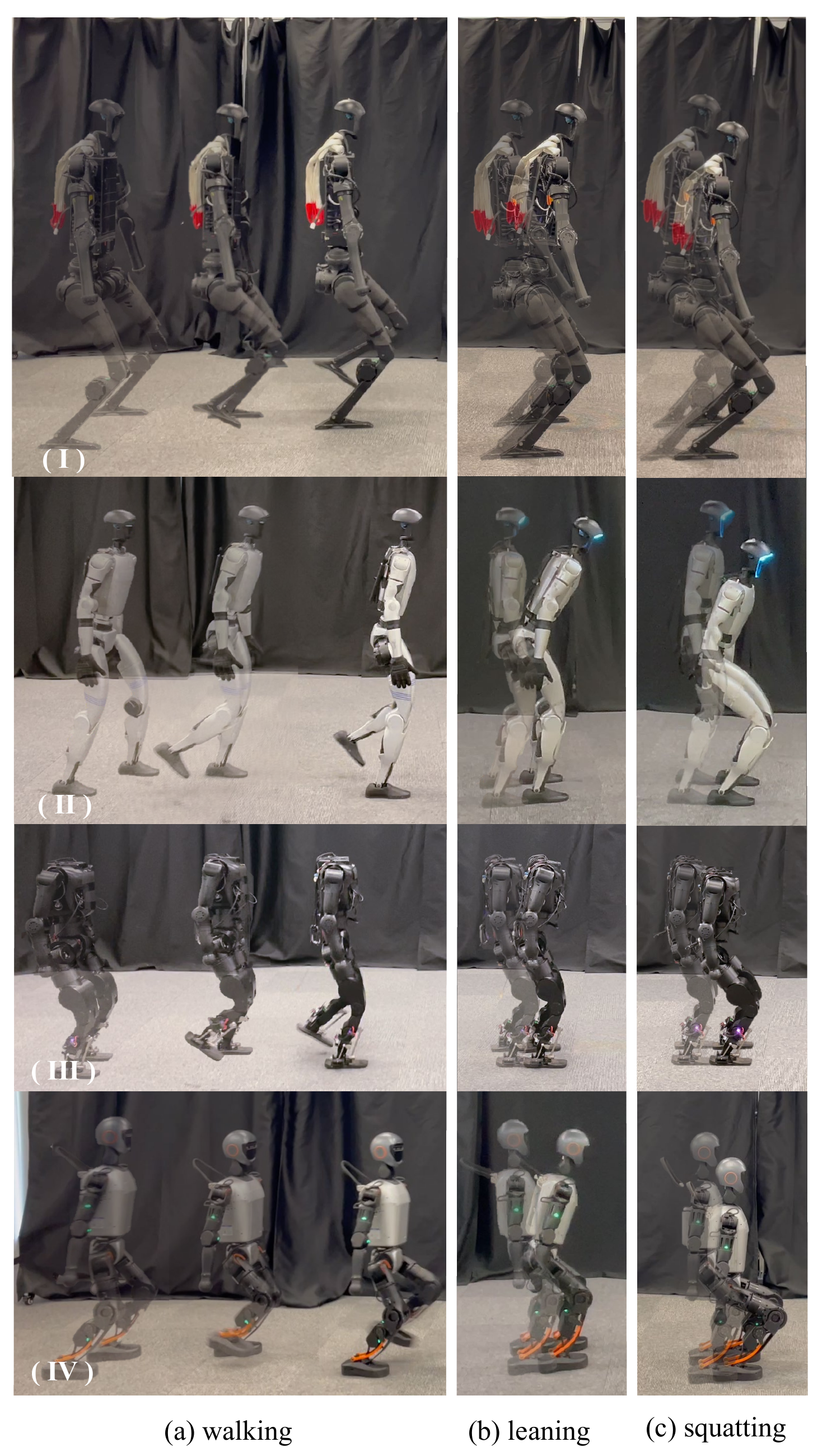} 
  \caption{\textbf{Real-world experiment.} We evaluate our policy across four embodiments: (I) Unitree H1, (II) Unitree G1, (III) Fourier N1, and (IV) Booster T1. Each robot successfully executes diverse commands, including (a) walking, (b) leaning, and (c) squatting, demonstrating robust zero-shot transfer from simulation to the real world.}
  \label{fig:real-exp}
    \vspace{-8pt} 
\end{figure}

To demonstrate the policy's ability to deploy on real-world hardware, we tested it on four real humanoid robots: Unitree H1, Unitree G1, Booster T1, and Fourier N1 in a zero-shot manner. We use UniCon~\cite{lin2026unicon} as a unified control interface for operating different robot platforms.

In Fig.~\ref{fig:real-exp} and Fig.~\ref{fig:real-exp2}, the policy, trained purely in simulation, executes diverse behaviors including walking, leaning, and squatting. Despite variations in morphology and hardware dynamics, the controller consistently produces stable motions across embodiments. These results confirm that our distillation framework supports robust sim2real transfer and enables a single policy to generalize effectively to heterogeneous humanoid platforms.

\section{Discussion}
In this work, we do not exhaustively evaluate on 
\emph{unseen} humanoid embodiments at test time. A promising direction is to
combine our distillation framework with explicit URDF randomization or
morphology randomization during training, as explored by previous works like~\cite{bohlinger2024one, liu2025locoformer,ai2025towards}, so the generalist learns to cover a wider space of structures. 

Another limitation is that our
embodiment-aware observation is still coarse; future work could enrich it with
finer-grained morphology descriptors such as limb lengths, joint topology, and
kinematic tree structure, which may further improve cross-embodiment
generalization.

\section{Conclusion}

In this paper, we present \textsc{EAGLE}, an embodiment-aware generalist-specialist distillation framework for unified humanoid whole-body control. By combining a high-dimensional command interface with an iterative distillation loop, \textsc{EAGLE} enables a single policy to control five structurally diverse humanoids without per-robot reward tuning. Through extensive simulation and real-world experiments, we show that our approach not only achieves superior command-tracking accuracy but also supports rich behaviors like squatting and leaning. These results highlight the potential of embodiment-aware distillation as a scalable solution for cross-embodiment learning in complex locomotion tasks.

\section{Acknowledgment}
The SJTU team is supported by National Natural Science Foundation of China (62322603), Shanghai Municipal Science and Technology Major Project (2025SHZDZX025D08) and Shanghai Artificial Intelligence Laboratory.

{\footnotesize
\bibliographystyle{IEEEtranN} 
\bibliography{references}
}



\end{document}